# SensitiveNets: Learning Agnostic Representations with Application to Face Images

Aythami Morales*, Julian Fierrez*, *Member, IEEE*, Ruben Vera-Rodriguez, Ruben Tolosana

**Abstract**— This work proposes a novel privacy-preserving neural network feature representation to suppress the sensitive information of a learned space while maintaining the utility of the data. The new international regulation for personal data protection forces data controllers to guarantee privacy and avoid discriminative hazards while managing sensitive data of users. In our approach, privacy and discrimination are related to each other. Instead of existing approaches aimed directly at fairness improvement, the proposed feature representation enforces the privacy of selected attributes. This way fairness is not the objective, but the result of a privacy-preserving learning method. This approach guarantees that sensitive information cannot be exploited by any agent who process the output of the model, ensuring both privacy and equality of opportunity. Our method is based on an adversarial regularizer that introduces a sensitive information removal function in the learning objective. The method is evaluated on three different primary tasks (identity, attractiveness, and smiling) and three publicly available benchmarks. In addition, we present a new face annotation dataset with balanced distribution between genders and ethnic origins. The experiments demonstrate that it is possible to improve the privacy and equality of opportunity while retaining competitive performance independently of the task.

—————————— ◆ ——————————

## 1 INTRODUCTION

DURING the last decade, the accuracy has been the key concern for researchers developing automatic decision-making algorithms. Recent progress under that umbrella has made possible and practical automatic decision-making in quite challenging problems including Computer Vision, Speech Recognition and Natural Language Processing. However, the recognition accuracy is not the only aspect to attend when designing learning algorithms. Algorithms have an increasingly important role in decision-making in several processes involving humans [1]. These decisions have therefore growing effects in our lives, and there is an increasing need for developing machine learning methods that guarantee fairness in such decision-making [2][3][4][5][6].

Discrimination can be defined in this context as the unfair treatment of an individual because of his or her membership in a particular group, e.g. ethnic, gender, etc. Privacy and discrimination protection are deeply embedded in the normative framework that underlies various national and international regulations. As a prove of these concerns, in April 2018 the European Parliament adopted a set of laws aimed to regularize the collection, storage and use of personal information, the General Data Protection Regulation (GDPR). According to paragraph 71 of the GDPR, data controllers who process sensitive data have to "implement appropriate technical and organizational measures …" that "… prevent, inter alia, discriminatory effects". GDPR prohibits any processing of user information with a purpose different of the originally declared [7]. Explicit information such as gender or ethnicity must be intentionally withhold of some automatic processes to avoid bias and discrimination. However, the last advances in machine learning allow to automatically extract sensitive information from unstructured data such as audio, text, and images [6][8]. Algorithms might intentionally or unintentionally exploit this information with undesirable discriminatory effects [1]. The GDPR encourages to integrate privacy preserving methods in the technology when created. In this context, how can we ensure that an algorithm might not access to this protected information?

The aim of this work is to develop a new privacy-preserving representation capable of removing certain sensitive information while maintaining the utility of the data. The proposed method, called SensitiveNets, can be trained for specific tasks (e.g. image classification), while minimizing the presence of selected covariates, both for the task at hand and in the information embedded in the trained network. These agnostic representations are expected to: i) improve the privacy of the data and the automatic process itself [8][9]; and ii) eliminate the source of discrimination that we want to prevent [10][11].

In particular, we evaluate the potential of SensitiveNets through the removal of the gender and ethnicity information from the embeddings of state-of-the-art face recognition systems. The proposed representation is evaluated on face images because of: i) the high level of sensitive information present in face imaging (e.g. gender, age, ethnic-ity, health) [12][13]; and ii) it is a challenging pattern recognition problem with multiple sources of variations (e.g. pose, illumination, image quality [13]).

The main contributions of this work: i) a new feature representation aimed at generating a learned embedding space that eliminates sensitive information from existing representations (Sect. 2); and ii) a new annotation dataset (DiveFace) made public in GitHub with uniform distribution between genders and ethnic origins (Sect. 3). The dataset includes more than 120K images from 24K identities.

After incorporating privacy into the learned space with SensitiveNets, we demonstrate in our experiments that sensitive attributes cannot be exploited in subsequent processes. SensitiveNets ensure both privacy-preserving embeddings (Sect. 4.3) and equality of opportunity of decision-making algorithms based on such embeddings (Sect. 4.4). The new SensitiveNets representation is achieved as a transform of a pre-trained feature space, being therefore compatible with existing pre-trained models. To the best of our knowledge, this is the first work that addresses this challenge for face recognition algorithms.

## 1.1 Related Works

The study of discrimination-aware information technology is not new and includes efforts from different research communities. In [15] researchers analyzed several techniques to improve fairness through discrimination-aware data mining. Similarly, a modified Bayes classifier focused on reducing discriminatory effects was proposed in [16], where the probability distributions of the classifiers were modified to guarantee fair decisions. Those approaches developed methods to act on the decisions rather than the learning processes.

On the other hand, researchers have also explored new fair representations capable of compensating unfair outcomes [3][4][17]. In [3][4] adversarial learning was used to improve three fairness criteria (demographic parity, equality of odds, and equality of opportunity). In [17] researchers proposed a gradient reversal training to improve fairness of the representations. The inclusion of fairness in the learning function allowed to reduce unfair outcomes in problems based on structured data [3][4][17]. However, the application of these approaches to train representations from unstructured data such as images was not developed.

Recent works have explored approaches to train fair representation in unstructured data such as images [6][10][11]. The proposal in [6] is based on a joint learning and unlearning algorithm inspired in domain and task adaptation methods. Similarly to [6], the authors of [11] propose a new regularization loss based on mutual information between feature embeddings and bias, training the networks using adversarial and gradient reversal techniques. The method in [10] was developed to train fair and more interpretable projections exploiting statistical differences between input data, interpretable projections, and the sensitive attributes.

Finally, privacy-preserving approaches have been proposed to disentangle certain attributes from learned representations. In [8][9] researchers proposed differential privacy approaches that obfuscate gender attributes at the image level while preserving face verification accuracy. These techniques generate realistic images capable of fooling human perception but fail in obfuscating the attributes at representation level (see Sect. 4.3). In [18][19] researchers proposed privacy-preserving techniques to disentangle variables of interest (e.g. facial expressions) from protected attributes (e.g. identity features). The methods, based on adversarial learning, reported encouraging privacy-preserving results, but at the cost of a non-negligible impact on the primary task performance

The methods proposed in [10][11] have been developed and evaluated for tasks involving a limited number of classes (e.g., digit classification, age prediction). As we will see in the Sect. 4.4, those approaches mitigate the bias but do not eliminate it. With SensitiveNets, instead of improving fairness like [3][4][17], we focus on improving the privacy of selected sensitive features. This way, fairness is not the objective, but the result of a privacy-preserving learning method capable of maintaining accuracies for the primary task.

## 2 PROPOSED METHOD

### 2.1. Problem formulation and framework

The feature vector $\mathbf{x} \in \mathbb{R}^d$ is a representation (also known as embedding) of an input sample $\mathbf{I_x}$ given a model with parameters $\mathbf{w} \in \mathbb{R}^M$. The model $\mathbf{w}$ is trained to obtain representations that maximize the inter-class distance and minimize the intra-class distance in a projected space (e.g., in face verification distance between faces from different and same identities, respectively).

The representation $\mathbf{x}$ is typically obtained as the output of one of the last layers of a trained deep neural network. Taking the top processing branch in Fig. 1, going from $\mathbf{x}$ to the final output of the trained deep network, the rest of the learning parameters are denoted as $\mathbf{w}_0$ (in our case a dense softmax layer with $C_k$ units). We suppose that the final output of the learning architecture is a vector of size $C_k$ containing the probabilities $\mathbf{p}_k(\mathbf{x})$ that $\mathbf{I_x}$ belongs to each of the classes of the task $k$.

In our framework, domain adaptation is used to learn new representations as transformations $\mathbf{f}_k(\mathbf{x})(k > 0)$ of the representation $\mathbf{x}$ learned originally for face recognition.

Without loss of generality, suppose that we have two of such transformations $\mathbf{f}_1$ and $\mathbf{f}_2$, which are trained specifically for a different task leaving fixed $\mathbf{w}$ as obtained in the learning architecture pre-trained for face recognition ($k = 0$). The learning process for a task $k > 0$ results in a vector of parameters $\mathbf{w}_k$ that describes both $\mathbf{f}_k(\mathbf{x})$ and the last dense softwax layer in that processing branch.

We propose to measure the information of the face em-beddings **x** generated by the pre-trained model **w** according to its performance in 3 different tasks: 1) Person Verification; 2) Gender Classification; 3) Ethnicity Classification.

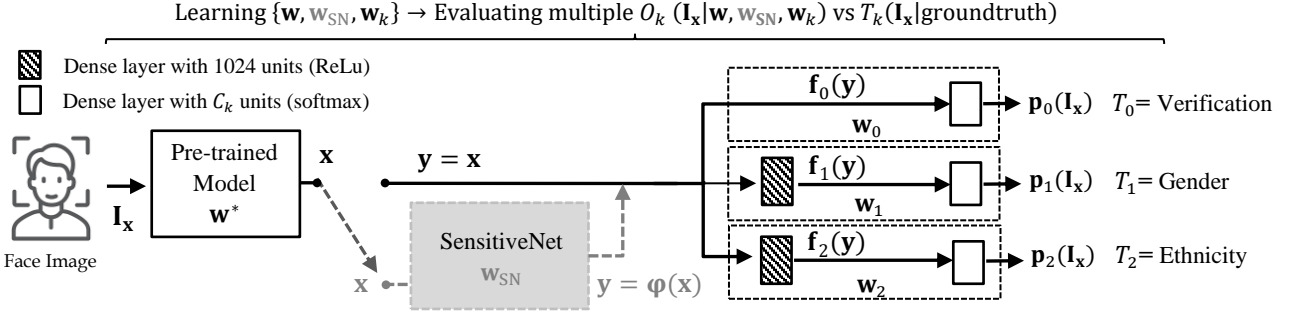

Fig. 1. Framework including domain adaptation from a pre-trained face representation **x** to multiple tasks (Verification, Gender, and Ethnicity classification) with and without the agnostic representation $\boldsymbol{\varphi}(\mathbf{x})$. $C_k$ is the number of classes for each task $k$ (e.g. $C_1 = 2$ corresponds to: *Male, Female*). $\mathbf{f}_k(\mathbf{x})$ is the projection for the adapted domain and $\mathbf{p}_k$ is the probability of $\mathbf{I_x}$ to belong to each of the classes of the task $k$.

The pre-trained model, represented by its parameters **w**, is trained for a given task $k$ (e.g. face verification, $k$=0 in Fig. 1) represented by a target function $T_k$, and a learning strategy that minimizes the error between an actual output $O_k$ of the full learning architecture and the target function $T_k$ (e.g. $T_0 = 1$ for matching face and $T_0 = 0$ for non-matching face). The learning strategy is traditionally based on the minimization of a loss function defined to obtain the best performance. The most popular approach for supervised learning in this setup is to train **w** and $\mathbf{w}_k$ by minimizing a loss function $\mathcal{L}_0$ over a set $\mathcal{E}$ of pre-training samples for which we have groundtruth targets:

$$\min_{\mathbf{w},\mathbf{w}_k} \sum_{\mathbf{I_x}\in\mathcal{E}} \mathcal{L}_0\big(O_k(\mathbf{I_x}|\mathbf{w},\mathbf{w}_k), T_k(\mathbf{I_x}|\text{groundtruth})\big) \quad (1)$$

As a result of the learning process, the solution $\{\mathbf{w}^*, \mathbf{w}_k^*\}$ to Eq. (1) generates a representation **x** that maximizes the discriminability of the feature space for the task $k$.

The goal of our proposed agnostic learning is to train a projection $\boldsymbol{\varphi}(\mathbf{x})$ (defined by its parameters $\mathbf{w}_{\text{SN}}$) that minimizes the performance of $\boldsymbol{\varphi}(\mathbf{x})$ for an specific task (e.g. $T_1$ or $T_2$ in Fig. 1), while maximizing it for other tasks (e.g. $T_0$). That objective can be achieved by solving (over a dataset $\mathcal{D}$ possibly different to $\mathcal{E}$):

$$\min_{\mathbf{w}_{\text{SN}}} \sum_{\mathbf{I_x}\in\mathcal{D}} \mathcal{L}_0\big(O_0(\mathbf{I_x}|\mathbf{w}^*, \mathbf{w}_{\text{SN}}, \mathbf{w}_0^*), T_0(\mathbf{I_x}|\text{groundtruth})\big) \\ + \mathcal{L}_k\big(O_k(\mathbf{I_x}|\mathbf{w}^*, \mathbf{w}_{\text{SN}}, \mathbf{w}_k^*), T_k(\mathbf{I_x}|\text{groundtruth})\big), \quad (2)$$

where $\mathcal{L}_k$ represents a loss function intended to minimize performance in the agnostic task $T_k (k > 0)$ while $\mathcal{L}_0$ tries to maximize performance in a different task $T_0$. This performance minimization for $T_k (k > 0)$ and maximization for $T_0$ can be interpreted as a kind of adversarial learning.

## 2.2. SensitiveNets: removing sensitive information

Triplet loss was originally proposed as a distance metric in the context of nearest neighborhood classification [20]. This distance was used to improve the performance of face descriptors in verification algorithms [21][22]. In this section we present SensitiveNets using triplet loss, but other loss functions can be used instead depending on the problem at hand with the methodology presented here (e.g., Sect. 4.4 uses binary cross-entropy loss).

Assume that each image is represented by an embedding descriptor $\mathbf{x} \in \mathbb{R}^d$ obtained by a pre-trained model $\mathbf{w}^*$. A triplet is composed by three different images from two different classes: Anchor (**A**) and Positive (**P**) are different images from the same class (e.g. an identity in face recognition), and Negative (**N**) is an image from a different class. We form a list $\mathcal{T}$ of triplets that satisfy:

$$\|\mathbf{x}_\mathbf{A}^i - \mathbf{x}_\mathbf{N}^i\|^2 - \|\mathbf{x}_\mathbf{A}^i - \mathbf{x}_\mathbf{P}^i\|^2 < \alpha, \quad (3)$$

where $i$ is the index of the triplet, $\|\cdot\|$ is the Euclidean Distance and $\alpha$ is a real numbered threshold. This list $\mathcal{T}$ includes a set of difficult triplets where the margin between the inter-class and the intra-class distances is limited by $\alpha$ as proposed in [21][22]. In our experiments $\alpha$ is equal to 0.2 and the number of triplets in $\mathcal{T}$ is around 100K.

Given the presented framework, SensitiveNets consists of: 1) assuming as input $\{\mathbf{w}^*, \mathbf{w}_0^*, \mathbf{w}_k^*\}$ (i.e., a pre-trained model $\mathbf{w}^*$, a task represented by $\mathbf{w}_0^*$ we aim to enforce, and a different task $k$ we aim to prevent), 2) activating the SensitiveNet block $\boldsymbol{\varphi}(\mathbf{x})$ in Fig. 1, and 3) solving the following version of Eq. (2):

$$\mathcal{L}_{\text{SN}} = \min_{\mathbf{w}_{\text{SN}}} \sum_{i\in\mathcal{T}} \Big[ \mathcal{L}'_0\big(\boldsymbol{\varphi}(\mathbf{x}_\mathbf{A}^i, \mathbf{x}_\mathbf{P}^i, \mathbf{x}_\mathbf{N}^i|\mathbf{w}_{\text{SN}})\big) + \\ + \Lambda_\mathbf{A}^i + \Lambda_\mathbf{P}^i + \Lambda_\mathbf{N}^i \Big], \quad (4)$$

where $\{\mathbf{x}_\mathbf{A}^i, \mathbf{x}_\mathbf{P}^i, \mathbf{x}_\mathbf{N}^i\}$ are the feature vectors of the triplet $i$ (note that a triplet by definition incorporates the groundtruth information indicated in Eq. (1)), $\mathcal{L}'_0$ is the triplet loss function of [20]:

$$\mathcal{L}'_0 = \|\boldsymbol{\varphi}(\mathbf{x}_\mathbf{A}^i) - \boldsymbol{\varphi}(\mathbf{x}_\mathbf{P}^i)\|^2 - \|\boldsymbol{\varphi}(\mathbf{x}_\mathbf{A}^i) - \boldsymbol{\varphi}(\mathbf{x}_\mathbf{N}^i)\|^2 + \alpha, \quad (5)$$

and $\Lambda^i$ is an adversarial sensitive regularizer used to measure the amount of sensitive information present in the learned model represented by $\mathbf{w}_{\text{SN}}$. $\Lambda^i$ is calculated as:

$$\Lambda^i(\mathbf{x}^i) = \log\big(1 + |0.9 - P_k(D^i|\boldsymbol{\varphi}(\mathbf{x}^i|\mathbf{w}^*, \mathbf{w}_{\text{SN}}), \mathbf{w}_k^*)|\big) \quad (6)$$

The probability $P_k$ of observing a fixed $D^i$ sensitive class (e.g. $D^i = Female$) in the face embedding after the sensitive information removal $\boldsymbol{\varphi}$ is initially obtained with the pre-trained gender and ethnicity classifiers ($\mathbf{w}_1^*$ and $\mathbf{w}_2^*$ are initially trained with **x**, and re-trained on $\boldsymbol{\varphi}(\mathbf{x})$ in each iteration), and then we iterate to solve Eq. (4). In Eq. (6) $|\cdot|$ is the absolute value, and $\Lambda$ will tend to zero for larger $P_k$. Therefore, by minimizing the $\Lambda$ terms in Eq. (4) we force the retraining of $\mathbf{w}_k^*$ to output the fixed demographic class $D^i$ for all images, in this way eliminating the capacity to detect other classes from the face representation $\boldsymbol{\varphi}(\mathbf{x})$. In other words, we unlearn the facial features necessary to differentiate between demographic classes.

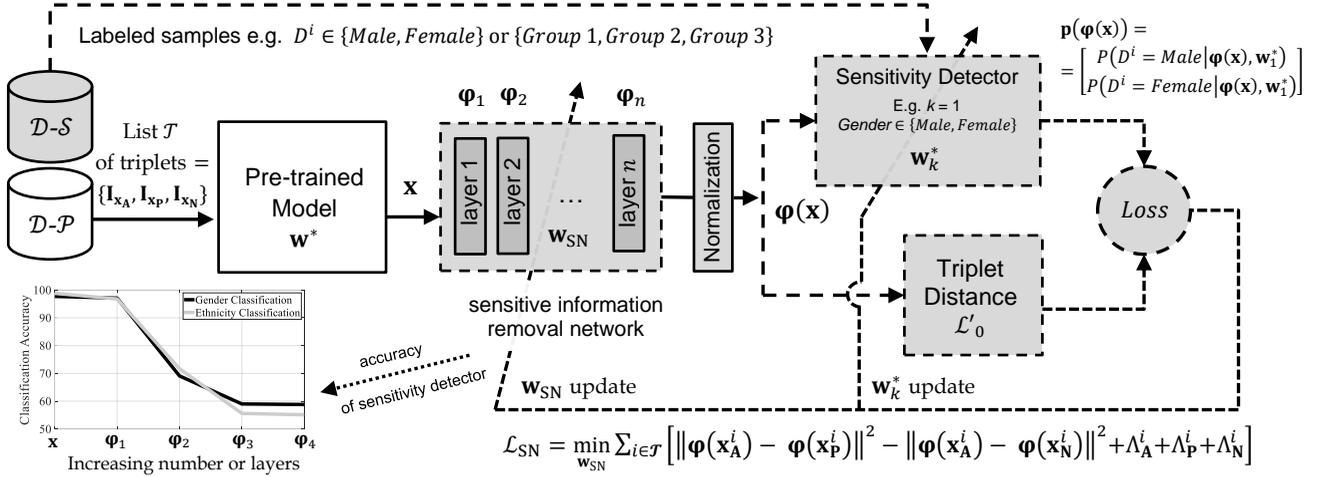

**Fig. 2.** Training process of SensitiveNets to remove sensitive information from the pre-trained embedding representation **x**. The Normalization is a $l_2$-norm and the Sensitivity Detector is trained using a softmax classification layer. The resulting feature representation is $\boldsymbol{\varphi}(\mathbf{x})$.

The network $\mathbf{w}_{SN}$ consists of $n$ dense layers with 1024 units each layer (linear activation). The layers are trained sequentially (from 1 to $n$) and each time a layer is trained, the sensitivity detectors $\mathbf{w}_1^*$ and $\mathbf{w}_2^*$ are re-trained to detect the sensitive information in the new learned representation $\boldsymbol{\varphi}$ using the data in $\mathcal{D}\text{-}\mathcal{S}$ (see Fig. 2). The redundancy in the feature space trained with Deep Neural Networks is usually very high. Sensitive information that was deprecated in the representation $\boldsymbol{\varphi}_j$ can be revealed and corrected in $\boldsymbol{\varphi}_{j+1}$ as we iteratively re-train $\mathbf{w}_k^*$. Note that we can eliminate multiple sensitive attributes as we train additional layers by including (or alternating) other tasks $\mathbf{w}_k^*$ anytime during training and fixing for them new labels $D^i$ in Eq. (6). In our experiments we remove in this way gender and ethnicity by alternating $D^i = Male$ and $D^i = ethnic\ Group\ 1$.

In Fig. 2 the update of $\mathbf{w}_k^*$ seeks to maximize the performance for task $k$ in each learning iteration. This is competing with the $\Lambda$ terms in Eq. (4), which aim at preventing the correct classification in that sensitive task. Overall, SensitiveNets as defined by Eqs. (4)-(6) and Fig. 2 can be interpreted as a kind of min-max adversarial formulation. Eq. (4) minimizes the sensitive information in $\boldsymbol{\varphi}(\mathbf{x})$ with the $\Lambda$ terms, trying to classify sensitive attributes based on $\boldsymbol{\varphi}(\mathbf{x})$ by updating $\mathbf{w}_k^*$ (with decreasing success as the learning progresses), and maintaining the performance in the primary task with the triplet loss term.

Note also that the training sets used must be labelled (i.e., targets $T_k$ available) for a Primary task we want to enforce ($k = 0$) and a Sensitive recognition task we want to prevent (e.g. $k = 1$ or $k = 2$), respectively, and both datasets ($\mathcal{D}\text{-}\mathcal{P}$ and $\mathcal{D}\text{-}\mathcal{S}$ for the Primary and Sensitive tasks) can be different (see Fig. 2). This provides important practical benefits as the size of the labelled sensitive attributes dataset can be much smaller than the size of the labelled dataset available for the primary task (which is normally the case, e.g., for face recognition).

For the problem experimentally addressed here (i.e., face recognition using a gender and ethnicity agnostic representation based on state-of-the-art deep networks and datasets), we have observed that it is necessary at least $n=3$ layers to obtain agnostic models.

## 3 DIVEFACE: DATASET FOR DIVERSITY-AWARE FACE RECOGNITION

An analysis of the 12 most cited face databases in the literature showed that Caucasian people represent more than 77% of the subjects in these databases, while for example Asian people only represent 9% [23]. Biased databases imply a double penalty for underrepresented classes. On the one hand, models are trained according to non-representative diversity. On the other hand, accuracies are measured on privileged classes and overestimate the real performance over a diverse society. Recently, diverse and discrimination-aware face databases have been proposed [24][25]. These databases present equal distribution of subjects among four ethnicities (Caucasian, Indian, Black, and Asian). However, gender balance is not considered. Each database includes their own biases (e.g., age of participants in [24], high quality of images in [25]). The creation of new databases like the previous ones with controlled biases is important to foster discrimination-aware research in machine learning and AI at large.

The database presented in this work, named DiveFace, is generated using images from the publicly available Megaface dataset MF2 [26] comprising 4.7M faces from 672K identities. Recently, Megaface dataset was decommissioned and images are no longer distributed by the University of Washington. All images of MF2 were obtained from Flickr and present realistic variations of pose, illumination, age, expression, and quality.

DiveFace contains annotations equally distributed among six classes related to gender and three ethnic groups. Gender and ethnicity have been annotated following a semi-automatic process (supervised learning plus manual inspection). In total, there are 24K identities (4K per class). The total number of images is greater than 120K, with an average number of images per identity of 5.5 and a minimum number of 3. Identities are grouped according to their gender (male or female) and three categories related to ethnic physical characteristics:

- Group 1: people with ancestral origin in Japan, China, Korea, and other countries in that region.
- Group 2: people with ancestral origins in Sub-Saharan Africa, India, Bangladesh, Bhutan, among others.
- Group 3: people with ancestral origins from Europe, North-America, and Latin-America.

We are aware about the limitations of grouping all human ethnic origins into only 3 categories. According to different studies, there are more than 5K ethnic groups in the world. We made the division in these three big groups to maximize differences among classes. As we will show in the experimental section, automatic classification algorithms based on these three categories show performances up to 98% accuracy.

## 4 EXPERIMENTS

### 4.1 Pre-trained model and databases

The performance of face recognition technology has been boosted significantly by deep convolutional neuronal networks in the last decade [28]. On the other hand, face images reveal information not only about who we are but also about demographics like gender, ethnicity, and age. Researchers have proposed to exploit such auxiliary data of the users to improve face recognition [29][30]. These auxiliary data are also known as soft biometrics, which refer to those biometrics that can distinguish different groups of people but do not provide enough information to uniquely identify a person [31]. These soft attributes can be extracted with high accuracy using just one face picture [29][32].

In our experiments we employ the popular face recognition pre-trained model ResNet-50. This model has been tested on competitive evaluations and public benchmarks [33]. ResNet-50 is a convolutional neural network with 50 layers and 41M parameters initially proposed for general purpose image recognition tasks [34]. The main difference with traditional convolutional neural networks is the inclusion of residual connections to allow information skip layers and improve gradient flow.

Our experiments include a ResNet-50 model trained from scratch using VGGFace2 dataset [33]. The pre-trained model is used as embedding extractor. Those embeddings are then $l_2$-normalised to generate our input representation **x**. The similarity between two face descriptors is calculated as the Euclidean distance between them. The verification accuracy is obtained comparing the distances between positive matches (belonging to the same identity) with negative matches (belonging to different identities). Two face descriptors are assigned to the same identity if their distance is smaller than a threshold. The pre-trained model used in this work achieved a verification accuracy (test set from view 1 experimental protocol) of 98.4% on the LFW benchmark [35].

DiveFace is employed to train the method proposed in Sect. 2. In order to demonstrate the generalization capability of the method, we evaluate the verification results over another two popular face datasets: Labeled Faces in the Wild (LFW) [35] and CelebA [27]. LFW is a database for research on unconstrained face recognition. The database contains more than 13K images of faces collected from the web. We consider the aligned images from the test set provided with view 1 and its associated evaluation protocol. CelebA is a large-scale face attributes dataset with more than 200K celebrity images. While the gender attributes are provided together with the CelebA dataset, ethnicity was labeled according to a commercial ethnicity detection system. These three databases are composed of images acquired in the wild, with large pose variations, varying face expressions, image quality, illuminations, and background clutter, among other variations [28][36].

### 4.2. Sensitive information in face descriptors

The first experiment aims to demonstrate the high level of sensitive information that forms part in face descriptors of state-of-the-art recognition algorithms. Following the framework presented in Sect. 2.1 and using the pre-trained model described in Sect. 4.1, we trained a classification layer (*softmax* activation function) composed of two or three neurons (for gender or ethnicity respectively). We used 9,000 and 1,800 images from DiveFace dataset for training and testing respectively (separate images and identities in each dataset). We kept frozen the parameters of the pre-trained models to train only the parameters of the classification layer ($\mathbf{w}_1$ and $\mathbf{w}_2$ in Fig. 1). To demonstrate the high presence of sensitive information in the embeddings generated by the pre-trained model, we report in Fig. 3 the classification accuracies of the model while reducing the number of features. Implementation details: 150 epochs, Adam optimizer (learning rate=0.001, $\beta_1 = 0.9$, and $\beta_2 = 0.999$), and batch size of 128 samples.

The results in Fig. 3 show that it is possible to accurately classify both gender and ethnicity even with only 10% of the features from the pre-trained model. It is important to highlight that Resnet-50 was trained for face verification, not gender or ethnicity classification. Although this model was trained for person recognition, sensitive information

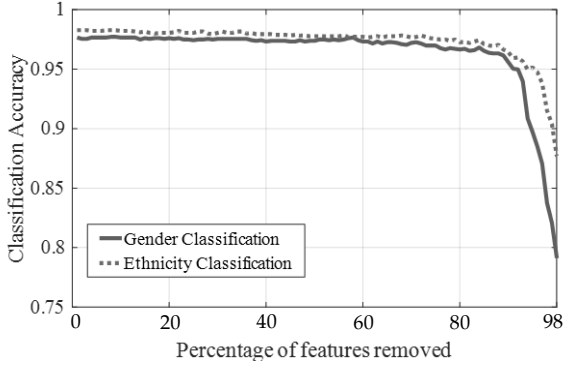

**Fig. 3.** Classification accuracy for gender and ethnicity vs percentage of features removed from the feature space before training.

TABLE I. CLASSIFICATION ACCURACIES FOR EACH TASK BEFORE AND AFTER APPLYING THE PROJECTION INTO THE NEW FEATURE REPRESENTATION. RECOGNITION REPRESENTS FACE VERIFICATION ACCURACY (IN %).

| Task | Before | After | Reduction* | Random |
|---|---|---|---|---|
| Recognition | 98.4% | 95.8% | 5.4% | 50% |
| Neural Network (NN) | | | | |
| Gender | 97.7% | 58.8% | 81.5% | 50% |
| Ethnicity | 98.8% | 55.1% | 66.4% | 33% |
| Support Vector Machine (SVM) | | | | |
| Gender | 96.2% | 56.3% | 86.4% | 50% |
| Ethnicity | 98.2% | 54.1% | 67.6% | 33% |
| Random Forest (RF) | | | | |
| Gender | 95.1% | 54.6% | 89.8% | 50% |
| Ethnicity | 97.3% | 53.5% | 68.1% | 33% |

*Reduction = (Before-After)/(Before-Random)

TABLE II. COMPARISON OF OUR METHOD TO THE GENDER DIFFERENTIAL PRIVACY METHOD IN [9] FOR REMOVING GENDER INFORMATION. GENDER CLASSIFICATION ACCURACIES FOR VARIOUS CLASSIFIERS (IN %).

| | Before | After Dif-Privacy [9] | After SensitiveNets |
|---|---|---|---|
| NN | 99.5% | 99.3% | 65.7% |
| SVM | 98.4% | 98.3% | 67.3% |
| RF | 98.5% | 98.5% | 65.2% |

is deeply embedded in its feature representation. According to these results, we can argue that sensitive features can be inferred from the embeddings. This may have a significant impact in the privacy of this sensitive information.

### 4.3 Removing sensitive information

The learning method proposed in Sect. 2 for obtaining the function $\boldsymbol{\varphi}(\mathbf{x})$ is trained using two different subsets of DiveFace. The sensitivity detector is trained with 3K different identities (3 images per identity) balanced between gender and ethnic groups. The list $\mathcal{T}$ of triplets is generated with the remaining 21K identities (all images available per identity) according to the Eq. (3) with $\alpha = 0.2$.

The aim of the proposed method is to maintain the face recognition performance while removing the sensitive information considered (gender and ethnicity). To analyze the effectiveness of the proposed method, we conducted two experiments including two datasets not used during the training phase of the agnostic features:

a) Maintaining performance on primary task: we calculated the face verification accuracy using either the original embeddings $\mathbf{x}$ or their projections $\boldsymbol{\varphi}(\mathbf{x})$ according to the evaluation protocol of the popular benchmark of LFW [35]. Table I shows the accuracies of embeddings generated by the pre-trained model before and after the proposed projection. The results show a small drop of performance when the projection is applied, which demonstrates the success of our method in preserving the accuracy in the main task here, i.e., face verification. Note that LFW was not used during the training process of SensitiveNet, and the high performance achieved demonstrates the capacity of the method to generalize to unseen databases.

b) Removing sensitive information: we train different gender and ethnicity classification algorithms (Neural Networks, Support Vector Machines, and Random Forests) either on original embeddings $\mathbf{x}$ or on their projections $\boldsymbol{\varphi}(\mathbf{x})$. The algorithms were trained and tested with 9,000 and 1,800 images, respectively. Table I shows the accuracies obtained by each classification algorithm before and after the projections. Results show a quite significant drop of performance in both gender and ethnicity classification when the proposed representation is applied, which demonstrates the success of our proposed approach in removing the sensitive information (gender and ethnicity in this case) from the embeddings.

We now apply a popular data visualization algorithm to gain insight about the presence of sensitive features in the embedding space generated by deep models. Fig. 4 (Left) shows the projection of each face into a 2D space generated from ResNet-50 embeddings using the t-SNE algorithm. After applying the unsupervised t-SNE 2D projection, we have colored each point according to its groundtruth ethnic and gender attributes. As we can see, the consequent face representation results in six clusters highly correlated with the demographic attributes. The gender and ethnicity information are highly embedded in the feature space and a simple t-SNE algorithm reveals the presence of this information. Fig. 4 (Right) shows the t-SNE projection of the same embeddings using $\boldsymbol{\varphi}(\mathbf{x})$. Note how the demographic clustering has disappeared for the learned representation $\boldsymbol{\varphi}(\mathbf{x})$ introduced in Sect. 2. These results suggest the potential of the proposed method to eliminate such demographic attributes from the face representations.

Table II shows the comparison between the proposed agnostic network and the gender differential privacy method in [9]. The authors of [9] provided a dataset composed by original and obfuscated versions of CelebA face images. ResNet-50 is used here to extract embeddings from both set of images. We trained three SVM classifiers using the embeddings from the original images (with and without SensitiveNets) and the obfuscated images. Table II shows the results. The differential-privacy approach is aimed at obfuscating the gender at image level, but fails in removing that information from the face descriptors at hand (when the gender detector $\mathbf{w}_1$ is trained using labels and obfuscated images). SensitiveNets reduces the performance of the gender classifier from 99% to 67%.

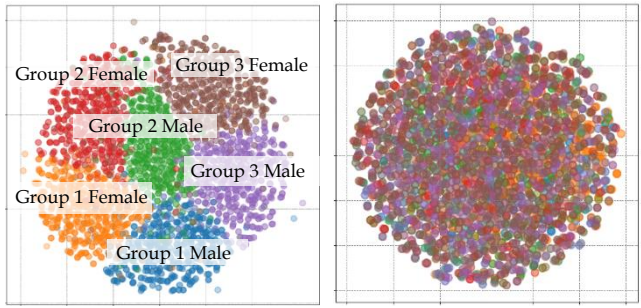

**Fig. 4.** Projections of the ResNet-50 embeddings **x** (Left) and **φ(x)** (Right) into the 2D space generated with t-SNE. (Color image)

## 4.4 Improving equality of opportunity

Inspired by the experiments carried out in [10][37], here we study how SensitiveNets representations can help to achieve a specific fairness criterion. We introduce two new tasks that we study separately as task number $k = 3$. This task number $k = 3$ is either binary Attractiveness classification or binary Smiling classification based on a face image $\mathbf{I_x}$. For this experiment, the method presented in Sect. 2.2 is trained to maintain the performance on the binary classifiers while eliminating the Gender information (task $k = 1$). To evaluate how the proposed method can generalize to other loss functions and tasks, the triplet loss function $\mathcal{L}'_0$ in Eq. (4) and (5) has been replaced by the popular *Binary Cross-Entropy*. The learned representation **φ(x)** is then used to train two binary SVM classifiers.

As fairness criterion, similar to [3][4][10] we use *Equality of Opportunity* [38]: the outcome of a binary classifier with input **x** and parameters $\mathbf{w}_3$ given its positive class should be independent to the feature $s$ we want to protect in terms of fairness. This criterion is particularly useful for classification problems where the positive class $T = 1$ is associated with an advantaged outcome.

Using the notation presented in Sect. 2.1 summarized in Fig. 1, this criterion results in: $\mathbf{p}_3(\mathbf{I_x}|\mathbf{w}^*, \mathbf{w}_3^*, T = 1, s) = \mathbf{p}_3(\mathbf{I_x}|\mathbf{w}^*, \mathbf{w}_3^*, T = 1)$, which implies equal True Positive Rates across the different possible values of $s$ for the trained Attractiveness or Smiling classifier characterized by $\mathbf{w}^*, \mathbf{w}_3^*$.

According to the method proposed in [10], we generated a gender biased training set where the proportion of attractive/smiling female and male subjects was 70% and 30% respectively (using CelebA dataset [27]). We introduced the opposite bias for the unattractive/not-smiling group with 30% and 70% of male and female, respectively. We also generated an unbiased evaluation dataset with 50% male and female subjects (randomly chosen). The experiment is performed using 40K images as training set, and 4K images for evaluation.

A classifier (SVM in our experiments) trained on face embeddings **x** generated by pre-trained models like ResNet-50, tends to reproduce the bias introduced in the training datasets. The results reported over the evaluation set in Table III show higher True Positive Rates (TPR) for the

TABLE III. RESULTS ON ATTRACTIVENESS/SMILING CLASSIFICATION. THE EQUAL OPPORTUNITY IS CALCULATED AS: TPR F - TPR M. F=FEMALE, M=MALE. BASELINE ACCURACIES IN BRACKETS.

| **Attractiveness** | Accuracy | TPR F | TPR M | Eq. Opp. |
|---|---|---|---|---|
| Fair [10]* | 79.4 (80.6) | 85.2 (90.8) | 61.4 (57.0) | **23.8 (33.8)** |
| LnL [11] | 73.4 (74.3) | 81.1 (92.6) | 68.3 (62.6) | **12.8 (30.1)** |
| SN [Ours] | 77.7 (74.3) | 81.4 (92.6) | 87.5 (62.6) | **6.8 (30.1)** |
| **Smiling** | | | | |
| LnL [11] | 87.3 (87.5) | 92.4 (93.8) | 83.5 (79.3) | **8.8 (14.5)** |
| SN [Ours] | 88.4 (87.5) | 90.9 (93.8) | 84.9 (79.3) | **6.0 (14.5)** |

*The accuracies in this case are directly extracted from [10].

privileged class (Female) in comparison with the non-privileged class (Male). In brackets, we show the baseline performance when training with the original representations. Table III shows how the agnostic representations **φ(x)** generated with SensitiveNets (SN in Table III) significantly reduce the gap between both classes (from 30.1% to 6.8% for Attractiveness and from 14.5 to 6.0 for Smiling classification). In addition, the overall accuracy is improved for both attributes. The agnostic representations avoid the network to exploit the latent variable related with the gender and reduce the impact of the biased training dataset. We also includes for comparison two other state-of-the-art methods proposed to unlearn protected attributes from face representations [10][11]. SensitiveNets outperforms (in term of equality of opportunity) the two other state-of-the-art methods (Fair and LnL in Table III) proposed for a similar objective: eliminating undesired information from learned representations. Note that while the method proposed in [11] was trained and evaluated using the same dataset that our method, the performance reported for the method proposed in [10] is the performance reported by the authors (using the same CelebA database but with a different split). Note also that in [11], the authors compared their method with previous approaches such as [6], showing a superior performance.

## 5 CONCLUSIONS

This work has proposed a privacy-preserving representation trained to eliminate sensitive information from deep neural network embeddings. The proposed representations are applicable to any machine learning problem and as a relevant example we have applied them to face images. Sensitive information such as gender or ethnicity is highly embedded in the feature space of most face descriptors, therefore face biometrics is an area particularly well suited for our methods.

The proposed agnostic representations are obtained by a new adversarial learning strategy called SensitiveNets, which maintains recognition performance while minimizing the presence of selected covariates. Our results show that it is possible to reduce the performance of gender and ethnicity detectors by 60-80% while the face verification performance is only reduced by 5%. The proposed SensitiveNets ensure both privacy-preserving embeddings (with respect to any sensitive feature we want to protect) and equality of opportunity of decision-making algorithms based on such embeddings. Recent applications of this method include facial gestures [39] or multimodal learning [40]. Additionally, we make available in GitHub a new annotation database (DiveFace) useful to train unbiased and discrimination-aware face recognition algorithms.

## ACKNOWLEDGMENT

This work has been supported by projects: PRIMA (MSCA-ITN-2019-860315), TRESPASS (MSCA-ITN-2019-860813), and BIBECA (RTI2018-101248-B-I00 MINECO).